\pdfoutput=1

\documentclass[11pt]{article}

\usepackage{EMNLP2022}
\usepackage{times}
\usepackage{latexsym}
\usepackage{amsmath}
\usepackage{multirow}
\usepackage{graphicx}
\usepackage{booktabs}
\usepackage{comment}
\usepackage{physics}
\usepackage{adjustbox}

\usepackage[T1]{fontenc}

\usepackage[utf8]{inputenc}

\usepackage{microtype}

\usepackage{inconsolata}

\usepackage{hyperref}       
\usepackage{url}            
\usepackage{amsfonts}       
\usepackage{nicefrac}       
\usepackage{microtype}      
\usepackage{xcolor}         

\title{GammaE: Gamma Embeddings for Logical Queries on Knowledge Graphs}

\author{Dong Yang$^{1}$\thanks{\ \ Equal contribution.} \thanks{\ \ Corresponding author (dongyang3-c@my.cityu.edu.hk).} , Peijun Qing$^{2\ast}$\thanks{\ \ The work was done during their internship in OPPO Research Institute.} , Yang Li$^{3\ddagger}$, Haonan Lu$^{1}$, Xiaodong Lin$^{4}$ \\
  $^{1}$OPPO Research Institute, Shenzhen, China \\
  $^{2}$Xidian University, Xi'an, China \\
  $^{3}$The Hong Kong Polytechnic University, Hong Kong, China \\
  $^{4}$Rutgers University, USA \\
  \texttt{\{yangdong1,luhaonan\}@oppo.com, qingpaine@gmail.com} \\
  \texttt{21041927g@connect.polyu.hk, lin@business.rutgers.edu} \\  
}

\begin{document}
\maketitle
\begin{abstract}
Embedding knowledge graphs (KGs) for multi-hop logical reasoning is a challenging problem due to massive and complicated structures in many KGs. Recently, many promising works projected entities and queries into a geometric space to efficiently find answers. However, it remains challenging to model the negation and union operator. The negation operator has no strict boundaries, which generates overlapped embeddings and leads to obtaining ambiguous answers. An additional limitation is that the union operator is non-closure, which undermines the model to handle a series of union operators. To address these problems, we propose a novel probabilistic embedding model, namely Gamma Embeddings (GammaE), for encoding entities and queries to answer different types of FOL queries on KGs. We utilize the linear property and strong boundary support of the Gamma distribution to capture more features of entities and queries, which dramatically reduces model uncertainty. Furthermore, GammaE implements the Gamma mixture method to design the closed union operator. The performance of GammaE is validated on three large logical query datasets. Experimental results show that GammaE significantly outperforms state-of-the-art models on public benchmarks. Our model is publicly available at \textcolor{blue}{\url{https://github.com/dyang67/GammaE}}.
\end{abstract}


\section{Introduction}

Most important advances encode knowledge into large-scale graph data to model real-world knowledge graphs (KGs), such as Wikidata \citep{abiteboul1995foundations,vrandevcic2014wikidata}, Microsoft Academic \cite{sinha2015overview}, Medical domain \cite{bodenreider2004unified,wishart2018drugbank}. A knowledge graph aims to represent entities and describe relations between concepts and entities. Multi-hop reasoning on KGs is a fundamental task of predicting answer entities of a graph query. Though previous methods have provided significant insight, this domain exists many challenging works due to complicated query structures and incomplete graph data. Since most real-world KGs consists of multiple unobserved edges and nodes, multi-hop reasoning cannot easily find the answering query. 

Knowledge graph reasoning can be represented by the first-order logic (FOL) queries with basic operators, such as the existential quantifier ($\exists$), conjunction ($\wedge$), disjunction ($\vee$), and negation ($\neg$). One regular set of such graph queries is the conjunctive query, which only consists of existential quantifiers ($\exists$) and conjunctions ($\wedge$) \cite{abiteboul1995foundations}. For example, a conjunctive query, such as "who are Nobel laureates from Germany?", is calculated on open-domain KGs. The laureates are existentially quantified variables in this query, where some laureates connect the Nobel prize to Germany. In terms of this task, we aim to predict potential answers that can involve missing edges. Besides, it is difficult to find all possible query structures since any graph query can be satisfied by many subgraphs.

Current methods \cite{hamilton2018embedding,zhang2021cone,ren2020query2box,jiangetal2019adaptive} address these problems by projecting entities and relations into a low-dimensional geometric space. These methods could quickly discover answers, since there is no need to track all intermediate entities. This research direction provides an effective solution for encoding semantic positions of entities by using their neighbor information. It is imperative that the solution is not required to track all intermediate entities, and only uses the nearest neighbor information in the geometric space to quickly get answers. However, geometric embeddings of the union operator cannot locate in a closed space, that is, the union of two embedding boxes is not a box or the same structure. Thus, FOL operations have difficulty in running the union operator \cite{ren2020beta,ren2020query2box}. Due to the non-smooth distance function, the distance between entities is difficult to calculate in the boxing space \cite{ren2020query2box}.

Other approaches apply the density function to encode entities and relations \cite{choudhary2021probabilistic,ren2020beta}. Though these models can solve all FOL queries, they cannot efficiently handle the union operator ($\vee$) due to its non-closure property in the density space. Besides, previous methods utilize the Disjunctive Normal Form (DNF) transformation which requires more computational steps. Thus, we aim to address the limitation and then implement the Gamma density function for encoding entities and relations. Another point is that our work can effectively process a stream of disjunctive queries.

Here we propose a Gamma Embedding (GammaE) probabilistic model to encode entities and relations for multi-hop reasoning on KGs. In the Gamma density space, all FOL operations, such as existential quantifier ($\exists$), intersection ($\wedge$), union ($\vee$), and negation ($\neg$), are closed and follow Gamma distributions. The linear property of the Gamma density can dramatically improve the computation efficiency for discovering answers. The contributions of our work are summarized as follows:

1. GammaE provides a closed solution for all FOL operators, including a projection, intersection, union, and negation. 

2. GammaE firstly implements the Gamma mixture method to alleviate the non-closure problem on union operators, which significantly reduces computation steps.

3. GammaE enables query embeddings to have strict boundaries on the negation operator, which can effectively avoid finding ambiguous answers.

4. GammaE outperforms state-of-the-art models on multi-hop reasoning over three benchmark datasets.

Our results have implications for encoding entities and relations, advancing the science of multi-hop reasoning, and improving our understanding of general knowledge graph. The rest of the paper is organized as follows: Section 2 shows the related work in multi-hop reasoning over KGs. Next, sections 3 and 4 theoretically demonstrate Gamma embeddings and define its FOL operations. The experimental setup and results are explicitly shown in section 5. Finally, section 6 makes a clear conclusion and section 7 briefly presents its limitations.

\section{Related Work}
This work is closely related to query embedding approaches \cite{hamilton2018embedding,daza2020message,vilnis2018probabilistic,alivanistos2021query,kotnis2021answering}. Most related models are applied for multi-hop reasoning over KGs \cite{ren2020query2box,zhang2021cone,choudhary2021probabilistic,ren2020beta,
arakelyan2020complex,ren2021smore,guo2018knowledge,lin2018multi,xiong2017deeppath,gu2015traversing}. Based on their embedding methods, these models can be categorized into two branches, namely geometric embedding models and probabilistic embedding models. The geometric embedding models assign adaptive geometric shapes for entities and relations with different structures \cite{ren2020query2box,boratko2021capacity,patel2021modeling}. These models map FOL queries into low-dimensional spaces and don't need to model the intermediate entities, which dramatically reduces the computation costs. They can effectively deal with a subset of FOL operations, while \citet{zhang2021cone} proposed a novel cone embedding for entities, and can handle all FOL operations. Meanwhile, \citet{bai2022query2particles} encode complex queries into multiple vectors, named particle embeddings. It also efficiently deals with all FOL operations. These models have strict borders relying on a non-smooth distance function, which difficultly handles multiple answers and easily ignores some correct entities. However, our model utilizes Gamma density to create linear space for each entity and relation, and build a soft-smooth distance function for alleviating the boundary effect.

Another method is to encode entities into the probabilistic density for performing multi-hop logical reasoning. \cite{chen2021probabilistic,li2018smoothing,dasgupta2020improving}. In vector embeddings, TransG \cite{xiao2015transg} firstly models the uncertainties of entities and relations by using the Gaussian mixture model. Their work doesn't be implemented into multi-hop reasoning on KGs. BETAE \cite{ren2020beta} firstly embeds entities and relations with the probabilistic density, which can handle all FOL operations. Since their union is non-closure, BETAE potentially leads to generate ambiguous entities between the two boxes. PERM \cite{choudhary2021probabilistic} applies the Gaussian density for representing entities and relations, and uses Gaussian mixture models for complex queries, i.e, long chains. However, PERM cannot handle the negation operator, and needs to spend large computational costs on learning knowledge graph representations. Our work firstly uses Gamma density to map entities and relations into linear space, and dramatically increases the performance and robustness. Also, GammaE efficiently finds the answers with closed and smooth forms.

\begin{figure*}[ht]
\vspace{-1.8em}
\centering

\includegraphics[width=16cm]{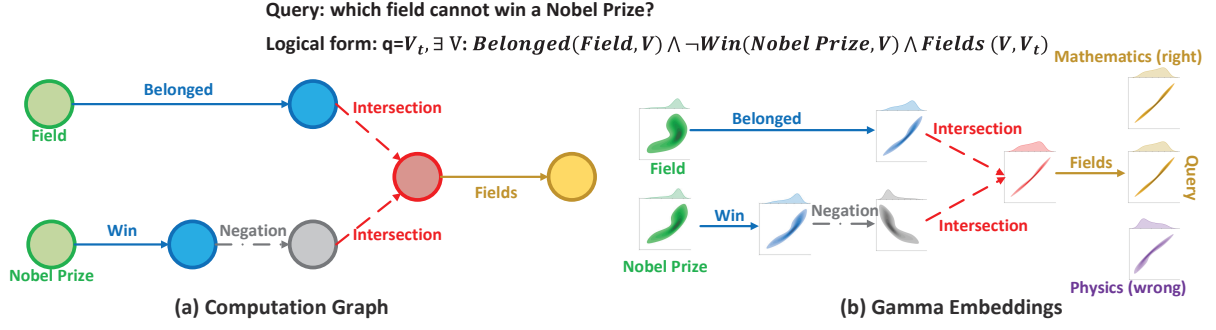}
\caption{Based on the computation graph, GammaE answers first-order logic queries including $\exists$, $\wedge$, and $\neg$ logical operators. (a) A specific query, such as "which field cannot win a Nobel Prize?", can be translated into its corresponding computation graph, where each node represents a set of entities and each edge stands for a logical operator. (b) GammaE encodes each node of the computation graph as a Gamma distribution over the entity embedding space. Each edge of the computation graph is calculated by logical operators, such as projection, negation, or intersection operators. The upper figure shows the probability distribution of the Gamma embedding, and the bottom figure reflects the contour mapping of the Gamma embedding. GammaE applies a series of logical operators to shape the size of the Gamma distribution. The answers of this query are then entities that are probabilistically close to the embedding of the query (e.g., the embedding of "Mathematics" is closer to the query embedding, not the embedding of "Physics").}
\label{fig1}
\vspace{-1.4em}
\end{figure*}

\section{Preliminaries}

A knowledge graph (KG) is a directed graph $G=(V,E,R)$, where $V$ is the set of entities, $E$ is the set of triplets, and $R$ denotes the set of relations. A direct triplet of a KG is represented as $(e_1,r,e_2) \in E$, that is, a relation $r$ linking the entity $e_1$ to the entity $e_2$, where $e_1,e_2 \in V$ and $r \in R$. For each triplet, it has a relational binary function, i.e, $r(e_1,e_2)={\rm True}$ if and only if $(e_1,r,e_2)$ exists. 

\textbf{First-order Logic Queries.} The first-order logic (FOL) queries contain four basic operators, namely existential quantifier ($\exists$), conjunction ($\wedge$), disjunction ($\vee$), and negation ($\neg$).
A first-order logic query $q$ consists of a non-variable anchor entity set $V_a \subseteq V$, existential quantified set $\{V_1,V_2,...,V_k\}$, and a target set $V_t$, i.e, query answers. The logical form of a disjunctive query $q$ can be written as
\begin{align} 
q[V_t]=V_t. \ \exists V_1,V_2,...,V_k : c_1 \vee c_2 \vee ... \vee c_n,
\end{align}
where $c_i=b_{i1} \wedge b_{i2} ... \wedge b_{im}$, and $b_{ij}=r(e_a,V)$ or $^{\neg} r(e_a,V)$ or $r(V',V)$ or $^{\neg}r(V',V)$, for $e_a \in V_a$, $V \in \{V_t, V_1,...,V_k\}$, $V' \in \{V_1,...,V_k\}$, and $V \neq V'$, $r \in R$. Here, $c_i$ is a conjunctive query with one or more literals $b_{ij}$. And $b_{ij}$ is an atomic formula or a negation. 

\textbf{Computation Graphs.} Each query can generate its corresponding computation graph, where entities are mapped into nodes, and relations with atomic formulas are calculated by logical operators. These logical operators are defined as follows:\begin{enumerate}

\item Relation Projection. A set of entities is $S \subseteq V$, a relation type is $r \in R$, and neighbors of entities $S'$ are defined as $\cup_{v \in S} A_r(v)$, where $A_r(v) \equiv \{v' \in N:r(v,v')={\rm True}\}$.

\item Intersection. Given sets of entities $\{S_1, S_2, ..., S_k\}$, their intersection is $\cap^{k}_{i=1}S_i$.

\item Union. For sets of entities $\{S_1, S_2, ..., S_k\}$, their union is $\cup^{k}_{i=1}S_i$.

\item Negation. A set of entities is $S \subseteq V$, and its complement is $\overline{S} \equiv V \backslash S$. 

\end{enumerate}

\section{Gamma Embeddings for Logical Reasoning}

To address the multi-hop reasoning on incomplete KGs, we propose a novel model GammaE, which encodes both entities and queries into Gamma distributions. Next, the related probabilistic logical operators are transformed into relation projection, intersection, union, and negation. GammaE provides an efficient method to handle arbitrary FOL queries. The schematics of GammaE answering graph queries are explicitly illustrated in Fig.~\ref{fig1}.

\subsection{Gamma Embedding for Entities and Relations}

A Gamma distribution is defined as
\begin{align} 
f(x; \alpha , \beta )=\frac{x^{ \alpha -1}e^{- \beta x} \beta^{\alpha}}{\Gamma(\alpha)},
\end{align}
where $x>0$, $\alpha>0$ is the shape, $\beta>0$ is the rate, and $\Gamma(\ast)$ is the Gamma function. Thus, the uncertainty of distribution can be obtained by information entropy:

\begin{align} H(X) =&E[-\ln f(x; \alpha , \beta )] \nonumber \\  
=&E[-\alpha \ln (\beta)+ \ln(\Gamma(\alpha)) \nonumber \\
&-(\alpha-1)\ln(X)+\beta X ]  \nonumber \\
=&\alpha- \ln(\beta)+ \ln(\Gamma(\alpha))+(1-\alpha)\psi(\alpha),
\end{align}
where $\psi(\ast)$ is the digamma function (see Appendix~\ref{sec:appendix11I}).

\subsection{Probabilistic Projection Operator}
Given a set of entities' embeddings $S$, the probabilistic projection operator maps from $S$ to another set $S'$ dependent on the relation type $r$. This operator could be defined
\begin{align} 
S'={\rm MLP_r}(S),
\end{align}
where $\rm MLP_r$ is a multi-layer perceptron network for given relation type $r$. The transformed set $S'$ is $\cup_{v \in S} A_r(v)$, where $A_r(v) \equiv \{v' \in N:r(v,v')={\rm True}\}$. It is essential that the projection operator represents a relation type $r$ from one set of entities to another fuzzy set. To avoid obtaining a huge number of answers, the Gamma embeddings are limited in a fixed size, scaling GammaE. A visualization of the projection operator is shown in Fig.~\ref{fig2}a.

\subsection{Probabilistic Intersection Operator}
For two input embeddings of two entities $S_1$, $S_2$, their intersection operator is defined as

\begin{align} 
P_{S_{{\rm Inter}}}=\frac{1}{Z} P^{w_1}_{S_1}P^{w_2}_{S_2},
\label{eq_21}
\end{align}
where $Z$ is a normalization constant, $P^{w_1}_{S_1}=f(x; w_1\alpha_1 , w_1\beta_1)$, $P^{w_2}_{S_2}=f(x; w_2\alpha_2 , w_2\beta_2)$, and $w_1+w_2$=1.
Since the product of Gamma distribution $f(x; \alpha , \beta )$ approximates to the linear summation of parameters ($\alpha, \beta$), Eq.~\ref{eq_21} can be derived

\begin{align} 
P_{S_{{\rm Inter}}} &\propto  x^{ \sum_{i=1}^{2} w_i(\alpha_i -1)}e^{\sum_{i=1}^{2}-w_i\beta_i x} \nonumber \\ 
          &\approx f(x; \sum_{i=1}^{2} w_i\alpha_i , \sum_{i=1}^{2} w_i\beta_i).
\label{eq_22}
\end{align}

Thus, given $k$ input embeddings {$S_1,S_2,...,S_k$}, the intersection of Gamma embeddings $P_{S_{Inter}}$ can be calculated as

\begin{align} 
P_{S_{{\rm Inter}}}=\frac{1}{Z} \prod_{i=1}^{k} P^{w_i}_{S_i},
\end{align}
where $Z$ is a normalization constant, $P^{w_i}_{S_i}=f(x; w_i\alpha_i , w_i\beta_i)$, and $\sum_{i=i}^{k} w_i =1$. A complete proof is presented in Appendix~\ref{sec:appendix11}. Fig.~\ref{fig2}b illustrates the intersection operation.

For learning the parameters $w_1,w_2,...,w_k$, we realize it with the self-attention mechanism. A single attention parameter is to

\begin{align} 
w_i=\frac{{\rm exp}({\rm MLP_{att}}(S_i))}{\sum_j {\rm exp}({\rm MLP_{att}}(S_j))}.
\end{align}

\begin{figure*}[ht]
\vspace{-1.4em}
\centering

\includegraphics[width=15cm]{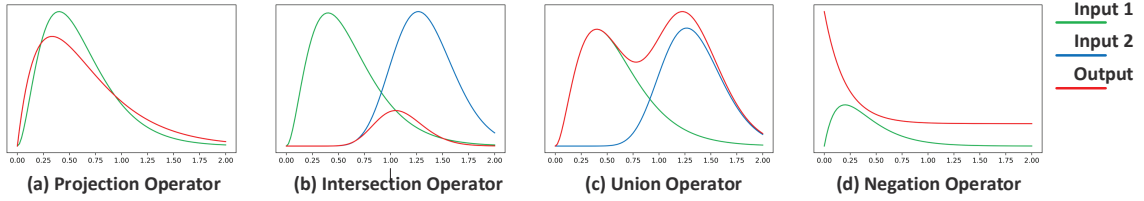}
\caption{The logical operators, such as projection, intersection, union, and negation, are demonstrated in the Gamma space. The operations are closed and will result in either a Gamma distribution or a Gamma mixture. The input embeddings are shown in \textcolor{green}{green} and \textcolor{blue}{blue}, and the corresponding output embeddings are represented in \textcolor{red}{red}.}
\label{fig2}
\vspace{-1.4em}
\end{figure*}

\subsection{Probabilistic Union Operator}
The union operator is implemented by Gamma mixture models. For $k$ input embeddings {$S_1,S_2,...,S_k$}, the union results can be calculated as

\begin{align} 
P_{S_{{\rm Union}}} &=\sum_{i=1}^{k} \theta_i P_{S_i},
\label{eq_23}
\end{align}
where $\theta_i=\frac{{\rm exp}(P_{S_i})}{\sum_j {\rm exp}(P_{S_i})}$, and $P_{S_i}=f(x;\alpha_i,\beta_i)$. Here, $\theta_i \in \Theta$ is the learned weight for each Gamma density in the Gamma mixture model and also uses the self-attention mechanism. Its operation is plotted in Fig.~\ref{fig2}c.

\subsection{Probabilistic Negation Operator}

A probabilistic negation operator takes Gamma embedding $S$ as the input, and then obtains an embedding of the complement $^{\neg}S (\overline{S})$. A desired property of the negation operator $N$ is to reverse in the sense where regions of high density in $P_{S}$ should have low probability density in $N_{P_{S}}$ and vice versa (Fig.~\ref{fig2}d). For Gamma embeddings, this negation operator for a given $S$ is defined as

\begin{align} 
N_{P_S} &=f(x;\frac{1}{\alpha},\beta)+\epsilon,
\label{eq_24}
\end{align}
where $P_S=f(x;\alpha,\beta)$ and $\epsilon \in (0,1)$ is the elasticity and set to $0.05$. The approach has one important advantage, that is, the pair of two Gamma embeddings has no intersection points. Furthermore, the elasticity $\epsilon$ can effectively increase the distance of two opposite embeddings. 

To avoid the identity problem in the negation, we design two labels ($0$ and $1$) to mark the original Gamma embedding ($0$) and its complement embedding ($1$). The label vector can effectively record this status for each entity.

\subsection{Learning Gamma Embeddings}

\textbf{Distance Function.} In our work, entities and queries are encoded into $m$-dimensional Gamma space. A entity $e$ is embedded by $P_e=[f(x;\alpha_1^e,\beta_1^e),...,f(x;\alpha_m^e,\beta_m^e)]$, and a query embedding $q$ is represented by $P_q=[f(x;\alpha_1^q,\beta_1^q),...,f(x;\alpha_m^q,\beta_m^q)]$. According to Kullback-Leibler (KL) divergence, the distance between two Gamma distributions is given by
\begin{align} 
&{\rm KL}(f(x;\alpha_e,\beta_e),f(x;\alpha_q,\beta_q))=(\alpha_e-\alpha_q)\psi(\alpha_e) \nonumber \\
&-\log \Gamma(\alpha_e)+\log \Gamma(\alpha_q)+\alpha_q(\log \beta_e-\log \beta_q) \nonumber \\
&+\alpha_e \frac{\beta_q-\beta_e}{\beta_e},
\label{eq_251}
\end{align}
where $\psi(\ast)$ is the digamma function. Its proof is shown in Appendix~\ref{sec:appendix211}.

Consequently, KL divergence of the entity $e$ and the query $q$ is obtained
\begin{align} 
dist(e;q) &=\sum_{i=1}^{m}{\rm KL}(P_e,i:P_q,i),
\label{eq_25}
\end{align}
where $P_e,i$ ($P_q,i$) represent the $i$-th Gamma distribution $f(x;\alpha_i^e,\beta_i^e)$ with parameters $\alpha_i^e$ and $\beta_i^e$ ($f(x;\alpha_i^q,\beta_i^q)$ with parameters $\alpha_i^q$ and $\beta_i^q$) in the entity (query) embedding vector. By this method, query embeddings will theoretically cover all answer entity embeddings \cite{kingma2013auto}.

\textbf{Training Objective.} In the training process, our objective is to minimize the KL divergence between Gamma embeddings of a query and its answer entity while maximizing the distance between that of this query and wrong answers via negative sampling method \cite{guo-etal-2016-jointly,xiong2017deeppath}. Thus, the loss objective is defined as

\begin{align} 
L=&-\log \sigma (\gamma-dist(v;q)) \nonumber \\
&-\sum_{j=1}^{k} \frac{1}{k} \sigma (dist(v'_j,q)- \gamma),
\label{eq_26}
\end{align}
where $v \in [q]$ represents the answer entity of $q$, $v'_j \notin |q|$ means a random negative sample, $k$ is the number of negative samples, $\gamma>0$ is the margin, and $\sigma(\ast)$ denotes the sigmoid function.  For inference, GammaE aims to find the answers of a query $q$, and ranks all the
entities based on the KL divergence defined in Eq.~\ref{eq_25} in constant time using Locality Sensitive Hashing \cite{indyk1998approximate}.

\begin{table*}[ht]
\vspace{-1.4em}
  \caption{MRR results (\%) on answering EPFO ($\exists$, $\wedge$, $\vee$) queries.}
  \centering
  \begin{tabular}{llllllllllll}
    \toprule
    \textbf{Dataset} &\textbf{Model} & \textbf{1p} & \textbf{2p} & \textbf{3p} & \textbf{2i} & \textbf{3i} & \textbf{pi} & \textbf{ip} & \textbf{2u} & \textbf{up} & \textbf{avg} \\
    \hline
    \multirow{5}{*}{FB15k} & GQE & 53.9& 15.5& 11.1& 40.2& 52.4& 27.5& 19.4& 22.3& 11.7& 28.2 \\
    		 & Q2B & 70.5& 23.0& 15.1& 61.2&71.8& 41.8& 28.7& 37.7& 19.0& 40.1  \\ 
    		 & BETAE& 65.1&25.7& 24.7& 55.8& 66.5& 43.9& 28.1& 40.1& 25.2& 41.6 \\
    		 & ConE& 73.3& 33.8& 29.2& 64.4& 73.7& 50.9& 35.7& 55.7& 31.4& 49.8 \\
    		 & GammaE& \pmb{76.5}& \pmb{36.9}& \pmb{31.4}& \pmb{65.4}& \pmb{75.1}& \pmb{53.9}& \pmb{39.7}& \pmb{57.1}& \pmb{34.5}& \pmb{52.3} \\
   	\hline
   	\multirow{5}{*}{FB15K-237} & GQE & 35.2& 7.4& 5.5& 23.6& 35.7& 16.7& 10.9& 8.4& 5.8& 16.6 \\
    		 & Q2B & 41.3&9.9&7.2&31.1&45.4&21.9&13.3&11.9&8.1&21.1   \\ 
    		 & BETAE& 39.0&10.9&10.0&28.8&42.5&22.4&12.6&12.4&9.7&20.9 \\
    		 & ConE& 41.8&12.8&11.0&32.6&47.3&25.5&14.0&14.5&10.8&23.4 \\
    		 & GammaE& \pmb{43.2}&\pmb{13.2}&\pmb{11.0}&\pmb{33.5}&\pmb{47.9}&\pmb{27.2}&\pmb{15.9}&\pmb{15.4}&\pmb{11.3}& \pmb{24.3}  \\
   	\hline
    \multirow{5}{*}{NELL995} & GQE & 33.1&12.1&9.9&27.3&35.1&18.5&14.5&8.5&9.0&18.7 \\
    		 & Q2B & 42.7&14.5&11.7&34.7&45.8&23.2&17.4&12.0&10.7&23.6  \\ 
    		 & BETAE& 53.0&13.0&11.4&37.6&47.5&24.1&14.3&12.2&8.5&24.6 \\
    		 & ConE& 53.1&16.1&13.9&40.0&50.8&26.3&17.5&15.3&11.3&27.2 \\
    		 & GammaE& \pmb{55.1}&\pmb{17.3}&\pmb{14.2}&\pmb{41.9}&\pmb{51.1}&\pmb{26.9}&\pmb{18.3}&\pmb{16.5}&\pmb{12.5}& \pmb{28.2}  \\	 
    \bottomrule
  \end{tabular}
  \label{table1}
\end{table*}

\begin{table*}[ht]
  \caption{HITS@1 results (\%) of GammaE, BETAE, Q2B, and GQE on answering EPFO ($\exists$, $\wedge$, $\vee$) queries.}
  \centering
  \begin{tabular}{llllllllllll}
    \toprule
    \textbf{Dataset} &\textbf{Model} & \textbf{1p} & \textbf{2p} & \textbf{3p} & \textbf{2i} & \textbf{3i} & \textbf{pi} & \textbf{ip} & \textbf{2u} & \textbf{up} & \textbf{avg} \\
    \hline
    \multirow{4}{*}{FB15k} & GQE & 34.2 & 8.3 & 5.0& 23.8& 34.9& 15.5&11.2& 11.5& 5.6& 16.6 \\
    		 & Q2B & 52.0& 12.7& 7.8& 40.5&53.4& 26.7& 16.7& 22.0& 9.4& 26.8  \\ 
    		 & BETAE& 52.0&17.0& 16.9& 43.5& 55.3& 32.3& 19.3& 28.1& 16.9& 31.3 \\
    		 & GammaE& \pmb{58.3}& \pmb{25.4}& \pmb{20.4}& \pmb{50.4}& \pmb{59.3}& \pmb{37.3}& \pmb{23.4}& \pmb{35.1}& \pmb{21.1}& \pmb{36.7} \\
   	\hline
   	\multirow{5}{*}{FB15K-237} & GQE &22.4 & 2.8& 2.1& 11.7& 20.9& 8.4& 5.7& 3.3& 2.1& 8.8 \\
    		 & Q2B & 28.3& 4.1 &3.0 &17.5&29.5&12.3&7.1&5.2&3.3&12.3   \\ 
    		 & BETAE& 28.9& 5.5 &4.9 &18.3&31.7&14.0&6.7&6.3&4.6&13.4 \\
    		 & GammaE& \pmb{32.1}&\pmb{9.7}&\pmb{8.3}&\pmb{22.3}&\pmb{35.6}&\pmb{18.3}&\pmb{8.7}&\pmb{9.6}&\pmb{8.3}& \pmb{16.6}  \\
   	\hline
    \multirow{5}{*}{NELL995} & GQE & 15.4&6.7&5.0&14.3&20.4&10.6&9.0&2.9&5.0&9.9 \\
    		 & Q2B & 23.8&8.7&6.9&20.3&31.5&14.3&10.7&5.0&6.0&14.1  \\ 
    		 & BETAE& 43.5&8.1&7.0&27.2&36.5&17.4&9.3&6.9&4.7&17.8\\
    		 & GammaE& \pmb{51.7}&\pmb{14.3}&\pmb{13.7}&\pmb{31.1}&\pmb{39.3}& \pmb{21.3}&\pmb{14.6}&\pmb{11.6}&\pmb{9.3}& \pmb{22.4}  \\	 
    \bottomrule
  \end{tabular}
  \label{tablec5}
\vspace{-1.4em}
\end{table*}

\section{Experiment}

GammaE is evaluated on three large-scale KG benchmark datasets, including FB15k \cite{bollacker2008freebase}, FB15k-237 \cite{toutanova2015observed}, and NELL995 \cite{xiong2017deeppath}. The experiment demonstrates GammaE on various tasks: 1. GammaE can efficiently handle all FOL queries. 2. GammaE outperforms the state-of-the-art baselines on the tasks. 3. The uncertainty of Gamma embeddings is well controlled. 4. All FOL operators are closed, especially the union and negation.

\textbf{Datasets.} For multi-hop reasoning, GammaE is studied on three standard benchmark datasets (details are shown in Appendix~\ref{sec:appendix31}):

\begin{itemize}
    \item FB15K \cite{bollacker2008freebase} contains the 149,689 entity pairs. Our experiment doesn’t contain invertible relations.
    \item FB15K-237 \cite{toutanova2015observed} consists of the 273,710 relation triples. All invertible relations are removed.
    \item NELL995 \cite{xiong2017deeppath} is collected by the 995th iteration of the Never-Ending Language Learning (NELL) system. It has 107,982 relation triples.
\end{itemize}

Based on previous experimental settings \cite{ren2020beta,ren2020query2box,hamilton2018embedding}, we evaluate GammaE on FOL queries without negation, and negation queries. GammaE focuses on answering queries involved with incomplete KGs. The training and testing queries consist of five conjunctive structures ($1p/2p/3p/2i/3i$) and five structures with negation ($2in/3in/inp/pni/pin$). Additionally, to evaluate the inference ability of the model, four extra query structures, namely $ip/pi/2u/up$, could show the ability to answer complicated queries with unseen structures during training. Please refer to Appendix~\ref{sec:appendix31} for more details about query structures.

\begin{table*}[ht]
  \caption{MRR results (\%) on answering graph queries with negation.}
  \centering
  \begin{tabular}{llllllll}
    \toprule
    \textbf{Dataset} &\textbf{Model} & \textbf{2in} & \textbf{3in} & \textbf{inp} & \textbf{pin} & \textbf{pni} & \textbf{avg} \\
    \hline
    \multirow{3}{*}{FB15k} & BETAE& 14.3&14.7& 11.5& 6.5& 12.4& 11.8 \\
    		 				& ConE& 17.9& 18.7& 12.5& 9.8& 15.1& 14.8\\
    		 				& GammaE& \pmb{20.1}& \pmb{20.5}& \pmb{13.5}& \pmb{11.8}& \pmb{17.1}& \pmb{16.6}\\
   	\hline
   	\multirow{3}{*}{FB15K-237} & BETAE & 5.1& 7.9& 7.4& 3.6& 3.4& 5.4\\
    		 & ConE & 5.4 &8.6&7.8&4.0&3.6&5.9   \\ 
    		 & GammaE& \pmb{6.7}&\pmb{9.4}&\pmb{8.6}&\pmb{4.8}&\pmb{4.4}&\pmb{6.78}\\
   	\hline
    \multirow{3}{*}{NELL995} & BETAE & 5.1&7.8&10.0&3.1&3.5&5.9 \\
    		 & ConE & 5.7&8.1&10.8&3.5&3.9&6.4  \\ 
    		 & GammaE& \pmb{6.3}&\pmb{8.7}&\pmb{11.4}&\pmb{4.0}&\pmb{4.5}&\pmb{6.98}\\	 
    \bottomrule
  \end{tabular}
  \label{table2}
\end{table*}

\begin{table*}[h]
  \caption{Spearman's rank correlation between learned embedding and the number of answers of queries.}
  \centering
  \begin{adjustbox}{width=\textwidth}
  \begin{tabular}{lllllllllllllll}
    \toprule
    \textbf{Dataset} &\textbf{Model} & \textbf{1p} & \textbf{2p} & \textbf{3p} & \textbf{2i} & \textbf{3i} & \textbf{pi} & \textbf{ip} & \textbf{2in} & \textbf{3in} & \textbf{inp} & \textbf{pin}& \textbf{pni} & \textbf{avg} \\
    \hline
    \multirow{3}{*}{FB15k} & BETAE & 0.37 & 0.48& 0.47& 0.57& 0.40& 0.52& 0.42& 0.62& 0.55& 0.46& 0.47& 0.61 & 0.50\\
    		 & ConE & 0.60& 0.68& 0.70& 0.68&0.52& 0.59& 0.56& \pmb{0.84}& 0.75& 0.61&0.58&\pmb{0.80} & 0.66\\ 
    		 & GammaE& \pmb{0.65}& \pmb{0.75}& \pmb{0.72}& \pmb{0.73}& \pmb{0.58}& \pmb{0.63}& \pmb{0.62}& 0.82& \pmb{0.79} &\pmb{0.65} & \pmb{0.63} & 0.77 & \pmb{0.70}\\
   	\hline
   	\multirow{3}{*}{FB15K-237} & BETAE & 0.42 & 0.55& 0.56& 0.59& 0.61& 0.60& 0.54& 0.71& 0.60 & 0.35& 0.45& 0.64 &0.55 \\
    		 & ConE & 0.56& 0.61& 0.60& \pmb{0.79}&0.79& 0.74& 0.58& \pmb{0.90}& 0.79& 0.56&0.48&\pmb{0.85} &0.69  \\ 
    		 & GammaE& \pmb{0.61}& \pmb{0.68}& \pmb{0.62}& 0.78& \pmb{0.83}& \pmb{0.79}& \pmb{0.63}&0.86 &\pmb{0.80}& \pmb{0.60} & \pmb{0.53} & 0.79& \pmb{0.71} \\
   	\hline
    \multirow{3}{*}{NELL995} & BETAE & 0.42 & 0.55& 0.56& 0.59& 0.61& 0.60& 0.54& 0.71& 0.60& 0.35& 0.45& 0.64 & 0.55 \\
    		 & ConE & 0.56& 0.61& 0.60& 0.79& 0.79& 0.74& 0.58& \pmb{0.90}& 0.79& 0.56&0.48&\pmb{0.85} & 0.69  \\ 
    		 & GammaE& \pmb{0.60}& \pmb{0.67}&\pmb{0.63} & \pmb{0.79}& \pmb{0.81}&\pmb{0.77} & \pmb{0.62}& 0.87 & \pmb{0.79}& \pmb{0.61} & \pmb{0.53} & 0.82 &\pmb{0.71}\\	 
    \bottomrule
  \end{tabular}
  \end{adjustbox}
  \label{table3}
\end{table*}
\textbf{Baseline.} We compare GammaE with state-of-the-art models, including GQE \cite{hamilton2018embedding}, Query2Box (Q2B) \cite{ren2020query2box}, BETAE \cite{ren2020beta}, and ConE \cite{zhang2021cone}. GQE and Q2B are trained only on five conjunctive structures ($1p/2p/3p/2i/3i$) as they cannot solve negation queries. For fair comparison, we assign the same dimensionality to the embeddings of these methods.

\textbf{Implementation Details.} In the training process, the weight $w$ and $\theta_i$ are calculated by the self-attention mechanism. For updating parameters, Adam is used as the optimizer \cite{kingma2014adam}. The hyperparameters are listed in Table~\ref{tablec311} (Appendix). For the evaluation, three KG datasets are built: the training KG dataset, the validation KG dataset, and the test KG dataset using training edges, training+validation edges, and training+validation+test edges, respectively. To reason over incomplete KGs, a given test (validation) query $q$ needs to discover non-trivial answers $[q]_{test} \textbackslash [q]_{val}$ ($[q]_{val} \textbackslash [q]_{train}$). To find answer entities, at least one edge is linked to one answer entity. For each non-trivial answer $v$ of a test query $q$, we rank it against non-answer entities $V \textbackslash [q]_{test}$. The evaluated metrics are selected by Mean Reciprocal Rank (MRR) and HITS@K, where a higher score means better performance. Their definitions are presented in Appendix~\ref{sec:appendix33}. All our models are implemented in Pytorch \cite{paszke2019pytorch} and run on one Tesla V100. The details of parameter settings are listed in Appendix~\ref{sec:appendix32}.

\subsection{Main Results}

In the experiments, we have run GammaE model 20 times with different random seeds, and reported the mean values of GammaE 's MRR results on EPFO and negation queries. For the error bars of main results, we report them in Appendix~\ref{sec:appendixebmr}. And the computational costs of GammaE are listed in Appendix~\ref{sec:appendixcct}.

\textbf{Modeling EPFO (containing only $\exists$, $\wedge$, and $\vee$) Queries.} First, we compare GammaE with baselines
that can only model queries with conjunction and disjunction without negation. Table 1 shows that GammaE achieves on average 5.0\%, 3.8\% and 3.7\% relative improvement MRR over previous state-of-the-art ConE on FB15k, FB15k-237, and NELL995, respectively. Previous works, like Q2B, BETAE, and ConE, use the disjunctive normal form (DNF) and De Morgan's laws (DM) to model the union operator, while we use more effective mixture methods to calculate the union operation. The average MRR of 2u/up is relatively improved by 5.5\% and 8.3\% over ConE on three benchmarks. Overall, GammaE on EPFO queries achieves better performance than five baselines over all three datasets.

Table~\ref{tablec5} shows the comparison of HITS@1 results for answering EPFO queries on three datasets. Compared to previous models (GQE, Q2B, and BETAE), GammaE significantly increases by 17.2\%, 23.9\%, and 25.8\% relative improvements over baselines on FB15k, FB15k-237, and NELL995, respectively.

\textbf{Modeling Queries with Negation ($\neg$).} Next, the performance of GammaE is evaluated to model queries with negation.  Since GQE and Q2B cannot handle the negation operator, they won't be compared in the experiments. Table~\ref{table2} shows results of GammaE outperform two baselines, including BETAE and ConE, on modeling FOL queries with negation. Specifically, GammaE achieves up to average 12.2\%, 14.9\%, and 9.1\% relative improvement MRR over two baselines on FB15k, FB15K-237, and NELL995, respectively. Another issue is that negation operators on BETAE and ConE exist boundary problems. The embedding of the negation output could have some intersections with the embedding of the input entity. However, the elasticity can effectively increase the distance between the input and the output during the negation operations.

\subsection{Modeling the Uncertainty of Queries}
To study the uncertainty of GammaE, we need to investigate the cardinality difference between a predicted set and an answer set. The cardinality can efficiently represent the uncertainty of an embedding model. For capturing the cardinality difference, we calculate the correlations between the differential entropy of the Gamma embedding $P_{[q]}$ and the cardinality of the answer set $[q]$. In this experiment, we choose Spearman's rank correlation coefficient (SRCC) and Pearson correlation coefficient (PCC) as the metric \cite{myers2004spearman,benesty2009pearson}. SRCC is designed to measure statistical dependence between the rankings of two variables, while PCC is a measure of linear correlation between two sets of data. Table~\ref{table3} shows SRCC comparisons of BETAE, ConE, and GammaE on three benchmark datasets. BETAE measures the uncertainty of the query by the differential entropy of Beta embeddings, while ConE calculates this uncertainty by the cone size. In Table~\ref{table3}, GammaE outperforms all previous models and relatively achieves up to 6.1\%, 2.9\%, and 2.9\% better correlation than ConE on B15k, FB15k-237, and NELL995. These results indicate that GammaE can effectively capture the uncertainty of a query. Besides, SRCC demonstrates Gamma embeddings can better model the cardinality of answer sets during training. PCC results are analyzed in Appendix~\ref{sec:appendixpcc}.

\subsection{Further Analysis of the Union Operator}
Existing models compute the union operator by De Morgan’s laws (DM) and disjunctive normal form (DNF). Due to De Morgan's laws, the union $\cup^{k}_{i=1}S_i$ can be rewritten as $\cup^{k}_{i=1}S_i=\overline{\cap ^{k}_{i=1}\overline{S_i}}$. The disjunctive normal form (DNF) is to move all "union" edges to the last step of the computation graph \cite{ren2020query2box}, which can reduce the number of training parameters. BETAE handles the union operator by DM since it can model the negation operator. BETAE also investigates the impact of DM and DNF on the union operator. Its results show that DNF is more robust than DM. According to this result, ConE adapts DNF to deal with the union operator. Though DNF and DM can compute the union operator, they are non-closure. Thus, to address this limitation, our work implements the Gamma mixture method to model the union operator. The Gamma mixture method can reduce computational steps. After passing the union operator, the output embeddings also follow the rule of the Gamma distribution. It is closed for modeling any queries with the union operator. In particular, Table~\ref{table1} shows that GammaE outperforms existing models for two types of union queries ($2u/up$), which indicates that the Gamma mixture method can effectively model the union operator. Additionally, Table~\ref{tablec1} further demonstrates the performance of GammaE with the mixture method (MM) is better than that of GammaE with DNF and DM.

\begin{table}[ht]
  \caption{MRR results (\%) for answering graph queries with union on FB15K, FB15K-237, and NELL995. The evaluated model is GammaE. Three union operators are DM, DNF, and Gamma mixture method, respectively. }
  \centering
  \begin{adjustbox}{width=\textwidth/2}
  \begin{tabular}{llll}
    \toprule
    \textbf{Dataset} &\textbf{Model} & \textbf{2u} & \textbf{up} \\
    \hline
    \multirow{3}{*}{FB15k} & GammaE with DM& 37.7&29.9 \\
    		 				& GammaE with DNF& 53.5& 30.9\\
    		 				& GammaE with MM& \pmb{57.1}& \pmb{34.5}\\
   	\hline
   	\multirow{3}{*}{FB15K-237} & GammaE with DM& 13.5& 10.1\\
    		 				& GammaE with DNF& 13.9& 10.3\\
    		 				& GammaE with MM& \pmb{15.4}& \pmb{11.3}\\
   	\hline
    \multirow{3}{*}{NELL995} & GammaE with DM& 14.5 & 10.9\\
    		 				& GammaE with DNF& 15.3& 11.2\\
    		 				& GammaE with MM& \pmb{16.5}& \pmb{12.5}\\	 
    \bottomrule
  \end{tabular}
  \end{adjustbox}
  \label{tablec1}
\end{table}

\subsection{Further Analysis of the Negation Operator}
\begin{table*}[ht]
  \caption{MRR results (\%) of GammaE with or without elasticity on answering graph queries with negation.}
  \centering
  \begin{tabular}{llllllll}
    \toprule
    \textbf{Dataset} &\textbf{Model} & \textbf{2in} & \textbf{3in} & \textbf{inp} & \textbf{pin} & \textbf{pni} & \textbf{avg} \\
    \hline
    \multirow{2}{*}{FB15k} & GammaE without elasticity $\epsilon$& 18.1&18.9&12.9&10.4&15.6&15.2\\
           & GammaE with elasticity $\epsilon$& \pmb{20.1}& \pmb{20.5}& \pmb{13.5}& \pmb{11.8}& \pmb{17.1}& \pmb{16.6}\\
    \hline
    \multirow{2}{*}{FB15K-237}
       & GammaE without elasticity $\epsilon$& 5.9&8.6&7.7&4.2&3.9&6.06\\
       & GammaE with elasticity $\epsilon$& \pmb{6.7}&\pmb{9.4}&\pmb{8.6}&\pmb{4.8}&\pmb{4.4}&\pmb{6.78}\\
    \hline
    \multirow{2}{*}{NELL995} & GammaE without elasticity $\epsilon$& 5.9&8.0&10.9&3.6&4.0&6.48\\ 
       & GammaE with elasticity $\epsilon$& \pmb{6.3}&\pmb{8.7}&\pmb{11.4}&\pmb{4.0}&\pmb{4.5}&\pmb{6.98}\\  
    \bottomrule
  \end{tabular}
  \label{tablec211}
\end{table*}
For the negation operator, BETAE takes the reciprocal of the parameter $\alpha$ and $\beta$, i.e., $N([(\alpha,\beta)]=[(\frac{1}{\alpha},\frac{1}{\beta})]$ \cite{ren2020beta}. ConE follows the rule of sector cones to model the negation operator \cite{zhang2021cone}. Their negation operators have the boundary problem, which could lead to obtaining ambiguous answers. The boundary effect means the original embedding has intersections with its complement embedding. In BETAE, the original embedding always has two crossover points with its complement embedding. By calculating the KL distance between two embeddings, the two crossover points definitely reduce the score. The above issues also exist in ConE as its original embedding shares same boundary edges with its complement embedding. However, GammaE effectively avoids the boundary effect due to using the elasticity.

One advantage of our negation operator is to design the elasticity. Since the negation operator aims to maximize the distance between the original embedding and its complement embedding, the elasticity could effectively increase the distance to obtain good performance. Table~\ref{tablec211} shows the elasticity $\epsilon$ can significantly improve the performance of GammaE.

Interestingly, since these queries are first-order queries, they only contain one negation operator. Therefore, our experiments didn't need to process two negation operators in a query, not facing identity cases.

\section{Conclusion}
In this paper, we propose a novel embedding model, namely GammaE, to handle arbitrary FOL queries and efficiently realize multi-hop reasoning on KGs. Given a query $q$, GammaE can map it onto the Gamma space for reasoning it by probabilistic logical operators on the computation graph. Compared to previous methods, its union operator uses the Gamma mixture model to avoid the disjunctive normal form and De Morgan's laws. Furthermore, GammaE significantly improves the performance of the negation operator due to alleviating the boundary effect. Extensive experimental results show that GammaE outperforms state-of-the-art models on multi-hop reasoning over arbitrary logical queries as well as modeling the uncertainty. Overall, GammaE aims to promote graph embeddings for logical queries on KGs.

\section{Limitations}
GammaE can handle all logical operators on large-scale KGs. Besides, all logical operators are closed in the Gamma space. It will significantly increase the capability and robustness of multi-hop reasoning on massive KGs. One potential risk is that the model could effectively model basic logical operators, not for more complicated operators or cycle graphs. If a query has many loops, the operations become harder. We will continue to work on this problem to design more effective logical operators. Importantly, we will continue to study this problem in the future.

\section*{Ethics Statement}
The authors declare that they have no conflicts of interest. This article does not contain any studies involving business data and personal information. 


\bibliography{emnlp2022}

\begin{thebibliography}{37}
\expandafter\ifx\csname natexlab\endcsname\relax\def\natexlab#1{#1}\fi

\bibitem[{Abiteboul et~al.(1995)Abiteboul, Hull, and
  Vianu}]{abiteboul1995foundations}
Serge Abiteboul, Richard Hull, and Victor Vianu. 1995.
\newblock \emph{Foundations of databases}, volume~8.
\newblock Addison-Wesley Reading.

\bibitem[{Alivanistos et~al.(2022)Alivanistos, Berrendorf, Cochez, and
  Galkin}]{alivanistos2021query}
Dimitrios Alivanistos, Max Berrendorf, Michael Cochez, and Mikhail Galkin.
  2022.
\newblock \href {https://openreview.net/forum?id=4rLw09TgRw9} {Query embedding
  on hyper-relational knowledge graphs}.
\newblock In \emph{International Conference on Learning Representations}.

\bibitem[{Arakelyan et~al.(2021)Arakelyan, Daza, Minervini, and
  Cochez}]{arakelyan2020complex}
Erik Arakelyan, Daniel Daza, Pasquale Minervini, and Michael Cochez. 2021.
\newblock \href {https://openreview.net/forum?id=Mos9F9kDwkz} {Complex query
  answering with neural link predictors}.
\newblock In \emph{International Conference on Learning Representations}.

\bibitem[{Bai et~al.(2022)Bai, Wang, Zhang, and Song}]{bai2022query2particles}
Jiaxin Bai, Zihao Wang, Hongming Zhang, and Yangqiu Song. 2022.
\newblock Query2particles: Knowledge graph reasoning with particle embeddings.
\newblock In \emph{2022 Annual Conference of the North American Chapter of the
  Association for Computational Linguistics}.

\bibitem[{Benesty et~al.(2009)Benesty, Chen, Huang, and
  Cohen}]{benesty2009pearson}
Jacob Benesty, Jingdong Chen, Yiteng Huang, and Israel Cohen. 2009.
\newblock Pearson correlation coefficient.
\newblock In \emph{Noise reduction in speech processing}, pages 1--4. Springer.

\bibitem[{Bodenreider(2004)}]{bodenreider2004unified}
Olivier Bodenreider. 2004.
\newblock The unified medical language system (umls): integrating biomedical
  terminology.
\newblock \emph{Nucleic acids research}, 32(suppl\_1):D267--D270.

\bibitem[{Bollacker et~al.(2008)Bollacker, Evans, Paritosh, Sturge, and
  Taylor}]{bollacker2008freebase}
Kurt Bollacker, Colin Evans, Praveen Paritosh, Tim Sturge, and Jamie Taylor.
  2008.
\newblock Freebase: a collaboratively created graph database for structuring
  human knowledge.
\newblock In \emph{Proceedings of the 2008 ACM SIGMOD international conference
  on Management of data}, pages 1247--1250.

\bibitem[{Boratko et~al.(2021)Boratko, Zhang, Monath, Vilnis, Clarkson, and
  McCallum}]{boratko2021capacity}
Michael Boratko, Dongxu Zhang, Nicholas Monath, Luke Vilnis, Kenneth~L
  Clarkson, and Andrew McCallum. 2021.
\newblock Capacity and bias of learned geometric embeddings for directed
  graphs.
\newblock \emph{Advances in Neural Information Processing Systems},
  34:16423--16436.

\bibitem[{Chen et~al.(2021)Chen, Boratko, Chen, Dasgupta, Li, and
  McCallum}]{chen2021probabilistic}
Xuelu Chen, Michael Boratko, Muhao Chen, Shib~Sankar Dasgupta, Xiang~Lorraine
  Li, and Andrew McCallum. 2021.
\newblock \href {https://doi.org/10.18653/v1/2021.naacl-main.68} {Probabilistic
  box embeddings for uncertain knowledge graph reasoning}.
\newblock In \emph{Proceedings of the 2021 Conference of the North American
  Chapter of the Association for Computational Linguistics: Human Language
  Technologies}, pages 882--893, Online. Association for Computational
  Linguistics.

\bibitem[{Choudhary et~al.(2021)Choudhary, Rao, Katariya, Subbian, and
  Reddy}]{choudhary2021probabilistic}
Nurendra Choudhary, Nikhil Rao, Sumeet Katariya, Karthik Subbian, and Chandan
  Reddy. 2021.
\newblock Probabilistic entity representation model for reasoning over
  knowledge graphs.
\newblock \emph{Advances in Neural Information Processing Systems}, 34.

\bibitem[{Dasgupta et~al.(2020)Dasgupta, Boratko, Zhang, Vilnis, Li, and
  McCallum}]{dasgupta2020improving}
Shib Dasgupta, Michael Boratko, Dongxu Zhang, Luke Vilnis, Xiang Li, and Andrew
  McCallum. 2020.
\newblock Improving local identifiability in probabilistic box embeddings.
\newblock \emph{Advances in Neural Information Processing Systems},
  33:182--192.

\bibitem[{Daza and Cochez(2020)}]{daza2020message}
Daniel Daza and Michael Cochez. 2020.
\newblock \href {https://arxiv.org/abs/2002.02406} {Message passing query
  embedding}.
\newblock In \emph{{ICML Workshop - Graph Representation Learning and Beyond}}.

\bibitem[{Guo et~al.(2016)Guo, Wang, Wang, Wang, and
  Guo}]{guo-etal-2016-jointly}
Shu Guo, Quan Wang, Lihong Wang, Bin Wang, and Li~Guo. 2016.
\newblock \href {https://doi.org/10.18653/v1/D16-1019} {Jointly embedding
  knowledge graphs and logical rules}.
\newblock In \emph{Proceedings of the 2016 Conference on Empirical Methods in
  Natural Language Processing (EMNLP)}, pages 192--202, Austin, Texas.
  Association for Computational Linguistics.

\bibitem[{Guo et~al.(2018)Guo, Wang, Wang, Wang, and Guo}]{guo2018knowledge}
Shu Guo, Quan Wang, Lihong Wang, Bin Wang, and Li~Guo. 2018.
\newblock Knowledge graph embedding with iterative guidance from soft rules.
\newblock In \emph{Proceedings of the AAAI Conference on Artificial
  Intelligence}, volume~32.

\bibitem[{Guu et~al.(2015)Guu, Miller, and Liang}]{gu2015traversing}
K.~Guu, J.~Miller, and P.~Liang. 2015.
\newblock Traversing knowledge graphs in vector space.
\newblock In \emph{Empirical Methods in Natural Language Processing (EMNLP)}.

\bibitem[{Hamilton et~al.(2018)Hamilton, Bajaj, Zitnik, Jurafsky, and
  Leskovec}]{hamilton2018embedding}
Will Hamilton, Payal Bajaj, Marinka Zitnik, Dan Jurafsky, and Jure Leskovec.
  2018.
\newblock Embedding logical queries on knowledge graphs.
\newblock \emph{Advances in neural information processing systems}, 31.

\bibitem[{Indyk and Motwani(1998)}]{indyk1998approximate}
Piotr Indyk and Rajeev Motwani. 1998.
\newblock Approximate nearest neighbors: towards removing the curse of
  dimensionality.
\newblock In \emph{Proceedings of the thirtieth annual ACM symposium on Theory
  of computing}, pages 604--613.

\bibitem[{Jiang et~al.(2019)Jiang, Wang, and Wang}]{jiangetal2019adaptive}
Xiaotian Jiang, Quan Wang, and Bin Wang. 2019.
\newblock \href {https://doi.org/10.18653/v1/N19-1103} {Adaptive convolution
  for multi-relational learning}.
\newblock In \emph{Proceedings of the 2019 Conference of the North {A}merican
  Chapter of the Association for Computational Linguistics: Human Language
  Technologies, Volume 1 (Long and Short Papers)}, pages 978--987, Minneapolis,
  Minnesota. Association for Computational Linguistics.

\bibitem[{Kingma and Ba(2015)}]{kingma2014adam}
Diederik~P. Kingma and Jimmy Ba. 2015.
\newblock \href {http://arxiv.org/abs/1412.6980} {Adam: A method for stochastic
  optimization}.
\newblock In \emph{International Conference on Learning Representations.
  (Poster)}.

\bibitem[{Kingma and Welling(2013)}]{kingma2013auto}
Diederik~P Kingma and Max Welling. 2013.
\newblock Auto-encoding variational bayes.
\newblock In \emph{International Conference on Learning Representations.}

\bibitem[{Kotnis et~al.(2021)Kotnis, Lawrence, and
  Niepert}]{kotnis2021answering}
Bhushan Kotnis, Carolin Lawrence, and Mathias Niepert. 2021.
\newblock Answering complex queries in knowledge graphs with bidirectional
  sequence encoders.
\newblock In \emph{Proceedings of the AAAI Conference on Artificial
  Intelligence}, volume~35, pages 4968--4977.

\bibitem[{Li et~al.(2018)Li, Vilnis, Zhang, Boratko, and
  McCallum}]{li2018smoothing}
Xiang Li, Luke Vilnis, Dongxu Zhang, Michael Boratko, and Andrew McCallum.
  2018.
\newblock Smoothing the geometry of probabilistic box embeddings.
\newblock In \emph{International Conference on Learning Representations}.

\bibitem[{Lin et~al.(2018)Lin, Socher, and Xiong}]{lin2018multi}
Xi~Victoria Lin, Richard Socher, and Caiming Xiong. 2018.
\newblock Multi-hop knowledge graph reasoning with reward shaping.
\newblock In \emph{Proceedings of the 2018 Conference on Empirical Methods in
  Natural Language Processing, {EMNLP}, Brussels, Belgium, October 31-November
  4, 2018}.

\bibitem[{Myers and Sirois(2004)}]{myers2004spearman}
Leann Myers and Maria~J Sirois. 2004.
\newblock Spearman correlation coefficients, differences between.
\newblock \emph{Encyclopedia of statistical sciences}, 12.

\bibitem[{Paszke et~al.(2019)Paszke, Gross, Massa, Lerer, Bradbury, Chanan,
  Killeen, Lin, Gimelshein, Antiga et~al.}]{paszke2019pytorch}
Adam Paszke, Sam Gross, Francisco Massa, Adam Lerer, James Bradbury, Gregory
  Chanan, Trevor Killeen, Zeming Lin, Natalia Gimelshein, Luca Antiga, et~al.
  2019.
\newblock Pytorch: An imperative style, high-performance deep learning library.
\newblock \emph{Advances in neural information processing systems}, 32.

\bibitem[{Patel et~al.(2021)Patel, Dangati, Lee, Boratko, and
  McCallum}]{patel2021modeling}
Dhruvesh Patel, Pavitra Dangati, Jay-Yoon Lee, Michael Boratko, and Andrew
  McCallum. 2021.
\newblock Modeling label space interactions in multi-label classification using
  box embeddings.
\newblock In \emph{International Conference on Learning Representations}.

\bibitem[{Ren et~al.(2021)Ren, Dai, Dai, Chen, Zhou, Leskovec, and
  Schuurmans}]{ren2021smore}
Hongyu Ren, Hanjun Dai, Bo~Dai, Xinyun Chen, Denny Zhou, Jure Leskovec, and
  Dale Schuurmans. 2021.
\newblock Smore: Knowledge graph completion and multi-hop reasoning in massive
  knowledge graphs.
\newblock \emph{arXiv preprint arXiv:2110.14890}.

\bibitem[{Ren et~al.(2020)Ren, Hu, and Leskovec}]{ren2020query2box}
Hongyu Ren, Weihua Hu, and Jure Leskovec. 2020.
\newblock Query2box: Reasoning over knowledge graphs in vector space using box
  embeddings.
\newblock In \emph{International Conference on Learning Representations}.

\bibitem[{Ren and Leskovec(2020)}]{ren2020beta}
Hongyu Ren and Jure Leskovec. 2020.
\newblock Beta embeddings for multi-hop logical reasoning in knowledge graphs.
\newblock \emph{Advances in Neural Information Processing Systems},
  33:19716--19726.

\bibitem[{Sinha et~al.(2015)Sinha, Shen, Song, Ma, Eide, Hsu, and
  Wang}]{sinha2015overview}
Arnab Sinha, Zhihong Shen, Yang Song, Hao Ma, Darrin Eide, Bo-June Hsu, and
  Kuansan Wang. 2015.
\newblock An overview of microsoft academic service (mas) and applications.
\newblock In \emph{Proceedings of the 24th international conference on world
  wide web}, pages 243--246.

\bibitem[{Toutanova and Chen(2015)}]{toutanova2015observed}
Kristina Toutanova and Danqi Chen. 2015.
\newblock Observed versus latent features for knowledge base and text
  inference.
\newblock In \emph{Proceedings of the 3rd workshop on continuous vector space
  models and their compositionality}, pages 57--66.

\bibitem[{Vilnis et~al.(2018)Vilnis, Li, Murty, and
  McCallum}]{vilnis2018probabilistic}
Luke Vilnis, Xiang Li, Shikhar Murty, and Andrew McCallum. 2018.
\newblock Probabilistic embedding of knowledge graphs with box lattice
  measures.
\newblock In \emph{Proceedings of the 56th Annual Meeting of the Association
  for Computational Linguistics (ACL: Long Papers)}.

\bibitem[{Vrande{\v{c}}i{\'c} and Kr{\"o}tzsch(2014)}]{vrandevcic2014wikidata}
Denny Vrande{\v{c}}i{\'c} and Markus Kr{\"o}tzsch. 2014.
\newblock Wikidata: a free collaborative knowledgebase.
\newblock \emph{Communications of the ACM}, 57(10):78--85.

\bibitem[{Wishart et~al.(2018)Wishart, Feunang, Guo, Lo, Marcu, Grant, Sajed,
  Johnson, Li, Sayeeda et~al.}]{wishart2018drugbank}
David~S Wishart, Yannick~D Feunang, An~C Guo, Elvis~J Lo, Ana Marcu, Jason~R
  Grant, Tanvir Sajed, Daniel Johnson, Carin Li, Zinat Sayeeda, et~al. 2018.
\newblock Drugbank 5.0: a major update to the drugbank database for 2018.
\newblock \emph{Nucleic acids research}, 46(D1):D1074--D1082.

\bibitem[{Xiao et~al.(2015)Xiao, Huang, Hao, and Zhu}]{xiao2015transg}
Han Xiao, Minlie Huang, Yu~Hao, and Xiaoyan Zhu. 2015.
\newblock Transg: A generative mixture model for knowledge graph embedding.
\newblock In \emph{Proceedings of the 54th Annual Meeting of the Association
  for Computational Linguistics (ACL: Long Papers)}.

\bibitem[{Xiong et~al.(2017)Xiong, Hoang, and Wang}]{xiong2017deeppath}
Wenhan Xiong, Thien Hoang, and William~Yang Wang. 2017.
\newblock Deeppath: A reinforcement learning method for knowledge graph
  reasoning.
\newblock In \emph{Proceedings of the 2017 Conference on Empirical Methods in
  Natural Language Processing (EMNLP)}.

\bibitem[{Zhang et~al.(2021)Zhang, Wang, Chen, Ji, and Wu}]{zhang2021cone}
Zhanqiu Zhang, Jie Wang, Jiajun Chen, Shuiwang Ji, and Feng Wu. 2021.
\newblock Cone: Cone embeddings for multi-hop reasoning over knowledge graphs.
\newblock \emph{Advances in Neural Information Processing Systems}, 34.

\end{thebibliography}
\bibliographystyle{acl_natbib}
\clearpage
\appendix
\section*{Appendix}
\setcounter{equation}{0}
\setcounter{subsection}{0}
\setcounter{table}{0}
\setcounter{figure}{0}
\renewcommand{\theequation}{A.\arabic{equation}}
\renewcommand{\thesubsection}{\Alph{section}.\arabic{subsection}}
\renewcommand\thetable{\Alph{section}\arabic{table}}   
\renewcommand\thefigure{\Alph{section}\arabic{figure}}  


The following is the supplementary Appendix for the paper. All the references are made in context of the main paper.

\section{Information Entropy of the Gamma Distribution}
\label{sec:appendix11I}
The information entropy of the Gamma distribution is defined as
\begin{equation}
\label{eq:gam-dent-s1}
\begin{split}
\mathrm{H}(X) =& \mathrm{E}[-\ln f(x; \alpha , \beta )]\\
=& - \mathrm{E}\left[ \ln \left( \frac{\beta^{\alpha}}{\Gamma(\alpha)} x^{\alpha-1} e^{-\beta x} \right) \right] \\
=& - \mathrm{E}\left[ \alpha \ln \beta - \ln \Gamma(\alpha) + (\alpha-1) \ln x - \beta x \right] \\
=& - \alpha \ln \beta + \ln \Gamma(\alpha) - (\alpha-1)\mathrm{E}(\ln x) \\
& + \beta\mathrm{E}(x) \; .
\end{split}
\end{equation}
Since the mean and expectation of the Gamma distribution
\begin{align}
\mathrm{E}(X) = \frac{\alpha}{\beta} \quad \text{and} \quad \mathrm{E}(\ln X) = \psi(a) - \ln(\beta),
\label{eq:gam-mean-logmean}
\end{align}
Eq.~\ref{eq:gam-dent-s1} can be solved

\begin{align}
\mathrm{H}(X) =&- \alpha  \ln \beta + \ln \Gamma(\alpha) - (\alpha-1)  (\psi(\alpha) \nonumber \\
& - \ln b) + \beta \frac{\alpha}{\beta} \nonumber \\
=& - \alpha  \ln \beta + \ln \Gamma(\alpha) + (1-\alpha)  \psi(\alpha) \nonumber \\
&+ \alpha \ln \beta - \ln \beta + \alpha \nonumber \\
=& \alpha - \ln \beta + \ln \Gamma(\alpha) + (1-\alpha) \psi(\alpha),
\label{eq:gam-dent-s2}
\end{align}
where $\psi(\ast)$ is the digamma function.

\section{Derivation of the Intersection Operator}
\label{sec:appendix11}
For two dimensions of gamma embeddings, the intersection operator is 

\begin{align}  \label{apeq1}
P_{S_{Inter}} =&\frac{1}{Z}P_{S_1}^{w_1}P_{S_2}^{w_2}   \nonumber  \\ 
     =&\frac{1}{Z}f(x;w_1 \alpha_1, w_1 \beta_1)f(x;w_2 \alpha_2, w_2 \beta_2) \nonumber \\ 
     =&\frac{1}{Z}\frac{x^{w_1 \alpha_1 -1}e^{- w_1 \beta_1 x}  (w_1 \beta_1)^{w_1 \alpha_1}}{\Gamma(w_1 \alpha_1)} \nonumber \\
     &\frac{x^{w_2 \alpha_2 -1}e^{- w_2 \beta_2 x}  (w_2 \beta_2)^{w_2 \alpha_2}}{\Gamma(w_2 \alpha_2)} \nonumber \\
=&\frac{(w_1 \beta_1)^{w_1 \alpha_1}(w_2 \beta_2)^{w_2 \alpha_2}}{Z\Gamma(w_1 \alpha_1)\Gamma(w_2 \alpha_2)} (x^{w_1 \alpha_1 -1}e^{- w_1 \beta_1 x})   \nonumber \\ 
    &(x^{w_2 \alpha_2 -1}e^{- w_2 \beta_2 x}) \nonumber \\
     =& \frac{(w_1 \beta_1)^{w_1 \alpha_1}(w_2 \beta_2)^{w_2 \alpha_2}}{Z\Gamma(w_1 \alpha_1)\Gamma(w_2 \alpha_2)} x^{\sum_{i=1}^{2}(w_i \alpha_i -1)} \nonumber \\
     &e^{- \sum_{i=1}^{2} w_i \beta_i x}.
\end{align}
Due to $w_1+w_2=1$, Eq.~\ref{apeq1} can be approximated as\begin{align} 
P_{S_{Inter}} =& \frac{(w_1 \beta_1)^{w_1 \alpha_1}(w_2 \beta_2)^{w_2 \alpha_2}}{Z\Gamma(w_1 \alpha_1)\Gamma(w_2 \alpha_2)} x^{\sum_{i=1}^{2}(w_i \alpha_i -1)}  \nonumber \\ 
 &e^{- \sum_{i=1}^{2} w_i \beta_i x} \nonumber \\
   \approx &  \frac{(w_1 \beta_1)^{w_1 \alpha_1}(w_2 \beta_2)^{w_2 \alpha_2}}{Z\Gamma(w_1 \alpha_1)\Gamma(w_2 \alpha_2)} x^{\sum_{i=1}^{2}w_i (\alpha_i -1)} \nonumber \\
   &e^{- \sum_{i=1}^{2} w_i \beta_i x} \nonumber \\
 =&Kx^{\sum_{i=1}^{2}w_i (\alpha_i -1)}e^{- \sum_{i=1}^{2} w_i \beta_i x} \nonumber \\
    \propto & x^{\sum_{i=1}^{2}w_i \alpha_i -1}e^{- \sum_{i=1}^{2} w_i \beta_i x} \nonumber \\
   \approx & f(x;\sum_{i=1}^{2}w_i \alpha_i, \sum_{i=1}^{2} w_i \beta_i),
\label{apeq2}
\end{align}
where $K= \frac{(w_1 \beta_1)^{w_1 \alpha_1}(w_2 \beta_2)^{w_2 \alpha_2}}{Z\Gamma(w_1 \alpha_1)\Gamma(w_2 \alpha_2)}$.

Based on Eq.~\ref{apeq2}, the intersection operator of $k$ Gamma embeddings can be obtained
\begin{align} 
P_{S_{Inter}} &=\frac{1}{Z}\prod_{i=1}^{k}P_{S_i}^{w_i} \nonumber  \\
   &\approx f(x;\sum_{i=1}^{k}w_i \alpha_i, \sum_{i=1}^{k} w_i \beta_i).
\label{apeq3}
\end{align}

\section{KL Divergence of Two Gamma Distributions}
\label{sec:appendix211}
One integral of two Gamma distributions has
\begin{footnotesize}
\begin{align}
&I(\alpha_e,\beta_e,\alpha_q,\beta_q) = \int_0^{\infty} \log (f(x;\alpha_e,\beta_e))f(x;\alpha_q,\beta_q) \dd x \nonumber  \\
  &= \int_0^{\infty} \log (\frac{x^{ \alpha_e -1}e^{- \beta_e x} \beta_e^{\alpha_e}}{\Gamma(\alpha_e)})\frac{x^{ \alpha_q -1}e^{- \beta_q x} \beta_q^{\alpha_q}}{\Gamma(\alpha_q)} \dd x \nonumber \\ 
&= -\beta_2 \int_0^{\infty} \frac{x^{ \alpha_q}e^{- \beta_q x} \beta_q^{\alpha_q}}{\Gamma(\alpha_q)} \dd x \nonumber \\
   &\quad - \log (\frac{\Gamma(\alpha_e)}{\beta_e^{\alpha_e}})\int_0^{\infty}\frac{x^{\alpha_q-1}e^{- \beta_q x} \beta_q^{\alpha_q}}{\Gamma(\alpha_q)} \dd x \nonumber \\
   &\quad +(\alpha_e-1) \int_0^{\infty} \log (x) \frac{x^{\alpha_q-1}e^{- \beta_q x} \beta_q^{\alpha_q}}{\Gamma(\alpha_q)} \dd x \nonumber \\
   &=-\frac{\alpha_q \beta_e}{\beta_q}-\log (\frac{\Gamma(\alpha_e)}{\beta_e^{\alpha_e}})+(\alpha_e-1) \nonumber \\
   &\quad \int_0^{\infty} \log (x) \frac{x^{\alpha_q-1}e^{- \beta_q x} \beta_q^{\alpha_q}}{\Gamma(\alpha_q)} \dd x.
\label{apeq4}
\end{align}
\end{footnotesize}

To solve the right integral in Eq.~\ref{apeq4}, one obtains
\begin{align} 
\pdv{\alpha_q} \Gamma(\alpha_q)=& \pdv{\alpha_q} \int_0^{\infty} x^{\alpha_q-1}e^{- \beta_q x} \beta_q^{\alpha_q} \dd x \nonumber \\
=& \pdv{(\alpha_q} \int_0^{\infty} e^{- \beta_q x} (x \beta_q)^{\alpha_q-1} \beta_q \dd x \nonumber \\
=& \int_0^{\infty} e^{- \beta_q x} x^{\alpha_q-1} \beta_q^{\alpha_q} \log x\beta_q \dd x \nonumber \\
=& \int_0^{\infty} \log(x)e^{- \beta_q x}x^{\alpha_q-1} \beta_q^{\alpha_q} \dd x \nonumber \\
& + \log(\beta_q)\Gamma(\alpha_q). 
\label{apeq5}
\end{align}

Since
\begin{align}
&\frac{\alpha_e-1}{\Gamma(\alpha_q)} \int_0^{\infty} \log(x)e^{- \beta_q x}x^{\alpha_q-1} \beta_q^{\alpha_q} \dd x \nonumber \\
&= (\alpha_e-1)\frac{\Gamma'(\alpha_q)}{\Gamma(\alpha_q)}-(\alpha_e-1)\log \beta_q,
\label{apeq6}
\end{align}


Eq.~\ref{apeq4} can be solved
\begin{align} 
I(\alpha_e,\beta_e,\alpha_q,\beta_q)=&-\frac{\alpha_q \beta_e}{\beta_q}-\log (\frac{\Gamma(\alpha_e)}{\beta_e^{\alpha_e}}) \nonumber \\
   &+(\alpha_e-1) \nonumber \\
   &\int_0^{\infty} \log (x) \frac{x^{\alpha_q-1}e^{- \beta_q x} \beta_q^{\alpha_q}}{\Gamma(\alpha_q)} \dd x \nonumber \\
   =&-\frac{\alpha_q \beta_e}{\beta_q}-\log (\frac{\Gamma(\alpha_e)}{\beta_e^{\alpha_e}}) \nonumber \\
    &+(\alpha_e-1)\frac{\Gamma'((\alpha_q)}{\Gamma(\alpha_q)}-(\alpha_e-1) \nonumber \\
    &\log \beta_q.
\label{apeq7}
\end{align}

Due to $\frac{\Gamma'(\alpha_q)}{\Gamma(\alpha_q)}=\psi(\alpha_q)$, Eq.~\ref{apeq7} can be rewritten as
\begin{align}
I(\alpha_e,\beta_e,\alpha_q,\beta_q)=&-\frac{\alpha_q \beta_e}{\beta_q}-\log (\frac{\Gamma(\alpha_e)}{\beta_e^{\alpha_e}}) \nonumber \\
    &+(\alpha_e-1)\psi(\alpha_q)-(\alpha_e-1) \nonumber \\
    &\log \beta_q.
\label{apeq8}
\end{align}
The KL divergence of $(f(x;\alpha_e,\beta_e))$ and $f(x;\alpha_q,\beta_q)$ can be obtained
\begin{align} 
&{\rm KL}(f(x;\alpha_e,\beta_e),f(x;\alpha_q,\beta_q))=\int_0^{\infty}f(x;\alpha_e,\beta_e) \nonumber \\
&\log\frac{f(x;\alpha_e,\beta_e)}{f(x;\alpha_q,\beta_q)} \nonumber \\
&=I(\alpha_e,\beta_e,\alpha_e,\beta_e)-I(\alpha_q,\beta_q, \alpha_e,\beta_e) \nonumber \\
&= (\alpha_e-\alpha_q)\psi(\alpha_e) -\log \Gamma(\alpha_e)+\log \Gamma(\alpha_q) \nonumber \\
&\quad+\alpha_q(\log \beta_e-\log \beta_q)+\alpha_e \frac{\beta_q-\beta_e}{\beta_e}.
\label{apeq9}
\end{align}

\begin{figure*}[!h]
\centering
\includegraphics[width=15cm]{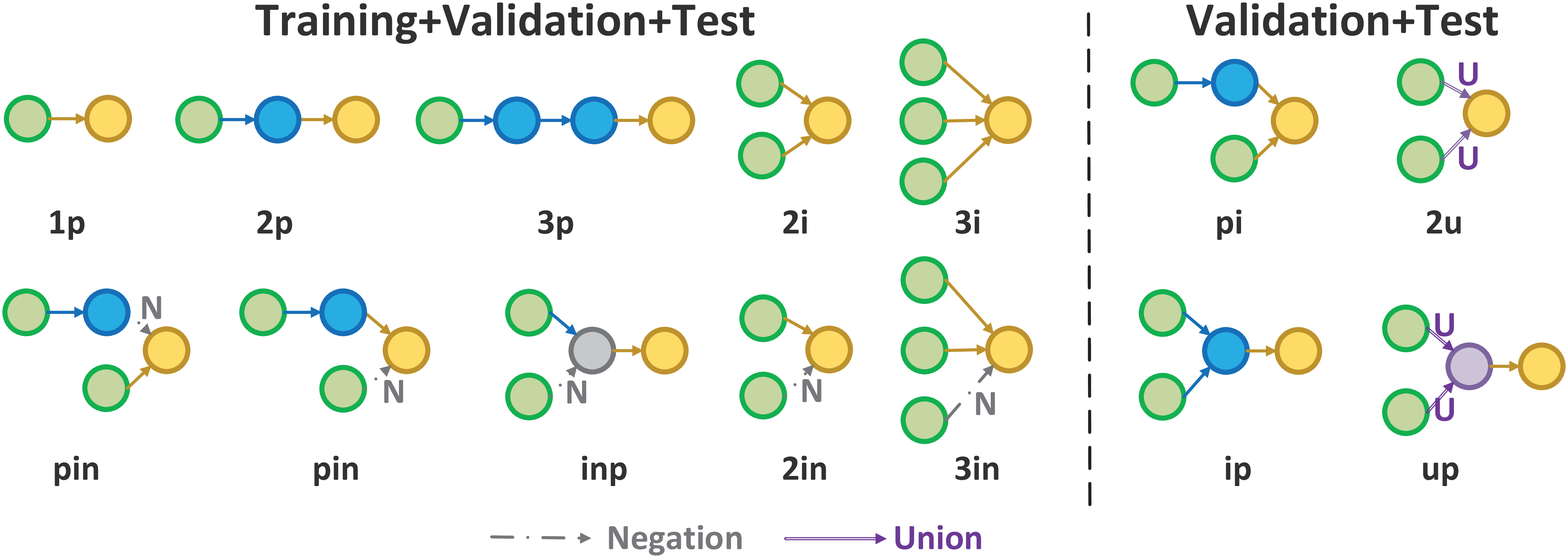}
\caption{Logical queries are illustrated by their computational graph structures. The left queries are used in the training phase, while all queries are evaluated in the validation and test phases. Note that $p$ means projection, $i$ is intersection, $n$ is negation, and $u$ is union.}
\label{fig3}
\end{figure*}

\section{Complementary Experimental Setup}
\label{sec:appendix3}
\subsection{Datasets and Query structures}
\label{sec:appendix31}
\begin{table*}[!ht]
  \caption{Statistics of query structures on three benchmark datasets, namely FB15K, FB15k-237, and NELL995.}
  \centering
  \begin{tabular}{l|cc|cc|cc}
    \toprule
    \textbf{Dataset} &\multicolumn{2}{c|}{\textbf{Training}} & \multicolumn{2}{c|}{\textbf{Validation}} & \multicolumn{2}{c}{\textbf{Test}} \\
    \hline
    \textbf{Dataset} &1p/2p/3p/2i/3i & negation & 1p & others & 1p & others  \\
    \hline
    FB15k & 273,710& 27,371&59,097 & 8,000 & 67,016 & 8,000\\   		 		
   	\hline
   	FB15K-237 & 149,689	& 14,968 & 20,101 & 5,000 & 22,812 & 5,000\\
   	\hline
    NELL995 & 107,982 & 10,798 & 16,927 & 4,000 & 17,034 & 4,000\\
    \bottomrule
  \end{tabular}
  \label{tablec2}
\end{table*}
Three datasets are used in the experiments, namely FB15k \cite{bollacker2008freebase}, FB15k-237 \cite{toutanova2015observed}, and NELL995 \cite{xiong2017deeppath}. For a fair comparison, fourteen query structures are created by the same rules in \citet{ren2020beta}. The related query structures are shown in Fig.~\ref{fig3}. These query structures fall into two categories, namely existential positive first-order (EPFO) queries and negative queries. EPFO queries consist of seven conjunctive structures ($1p/2p/3p/2i/3i/ip/pi$) and two disjunctive structures ($2u/up$). Negative queries contain five structures with negation ($2in/3in/inp/pni/pin$). The training dataset is composed of five conjunctive structures ($1p/2p/3p/2i/3i$) and five queries with negation ($2in/3in/inp/pni/pin$). The validation dataset and test dataset contain all logical structures, which never occurred during training. Table~\ref{tablec2} lists the basic statistics of different queries on three benchmark datasets.

\subsection{Parameter Settings}
\label{sec:appendix32}
\begin{table*}[!h]
  \caption{Hyperparameters of GammaE.}
  \centering
  \begin{adjustbox}{width=\textwidth}
  \begin{tabular}{lcccccc}
    \toprule
    \textbf{Dataset} &\textbf{embedding dim} & \textbf{learning rate} & \textbf{negative sample size $k$} & \textbf{margin $\gamma$} & \textbf{elasticity $\epsilon$} & \textbf{batch size} \\
    \hline
    FB15k & 800 & $5 \times 10^{-5}$ & 128 & 30 & 0.05 & 512 \\
    FB15K-237 & 800 & $1 \times 10^{-4}$ & 128 & 30 & 0.07 & 512 \\
    NELL995 & 800 & $1 \times 10^{-4}$ & 128 & 30 & 0.07 & 512 \\	 
    \bottomrule
  \end{tabular}
  \end{adjustbox}
  \label{tablec311}
\end{table*}
To obtain best results, we finetune these hyperparameters, such as embedding dimensions ($200,400,600,800$), the learning rate ($5 \times 10^{-5},10^{-4}, 5 \times 10^{-4},10^{-3}$), the negative sample size $k$ ($32,64,128$), the margin $\gamma$ ($20,30,40,50,60$), the elasticity $\epsilon$ ($10^{-3},0.01,0.03,0.05,0.07$), and the batch size ($128,256,512$). For all modules with MLP, we implement three-layer MLP and ReLU activation. The hyperparameters of GammaE are listed in Table~\ref{tablec311}.

\subsection{Evaluation Metrics}
\label{sec:appendix33}
In the main paper, Mean Reciprocal Rank (MRR) is used as the evaluation metric. Given multiple queries $Q$, MRR is the mean of the $Q$ reciprocal ranks, i.e.,
\begin{align} 
{\rm MRR}=\frac{1}{Q} \sum_{i=1}^{Q} \frac{1}{{\rm rank}_i},
\end{align}
where ${\rm rank}_i$ is the rank position of the first answer for the $i$-th query. 

HITS@K is also selected as the metric for evaluating our model. HITS@K is the rate of answers appearing in the top k entries for each instance list. It can be written as
\begin{align} 
{\rm HITS@K}=\frac{1}{K} \sum_{i=1}^{K} f({\rm rank}_i),
\end{align}
where $f({\rm rank}_i)$ is to 1 if ${\rm rank}_i \leq K$, otherwise, $f({\rm rank}_i)=0$.

\section{Complementary Experimental Results}
\label{sec:appendix1}

\subsection{Error Bars of Main Results}
\label{sec:appendixebmr}
\begin{table*}[ht]
\vspace{-1.4em}
  \caption{GammaE: MRR mean values and standard variances (\%) on answering EPFO ($\exists$, $\wedge$, $\vee$) queries.}
  \centering
  \begin{adjustbox}{width=\textwidth}
  \begin{tabular}{lcccccccccc}
    \toprule
    \textbf{Dataset} & \textbf{1p} & \textbf{2p} & \textbf{3p} & \textbf{2i} & \textbf{3i} & \textbf{pi} & \textbf{ip} & \textbf{2u} & \textbf{up} & \textbf{avg} \\
    \hline
    \multirow{2}{*}{FB15k} & 76.5& 36.9& 31.4& 65.4& 75.1& 53.9& 39.7& 57.1& 34.5&52.3\\
    	& \small \pmb{$\pm$ 0.081}& \small \pmb{$\pm$ 0.173}& \small \pmb{$\pm$ 0.169}& \small \pmb{$\pm$ 0.153}&\small \pmb{$\pm$ 0.198}&\small \pmb{$\pm$ 0.155}&\small \pmb{$\pm$ 0.132}&\small \pmb{$\pm$ 0.207}& \small \pmb{$\pm$ 0.185}&\small \pmb{$\pm$ 0.071} \\
   	\hline
   	\multirow{2}{*}{FB15K-237} & 43.2& 13.2&11.0&33.5&47.9&27.2&15.9&15.4&11.3&24.3\\
   	&\small \pmb{$\pm$ 0.061}&\small \pmb{$\pm$ 0.121}&\small \pmb{$\pm$ 0.153}&\small \pmb{$\pm$ 0.054}&\small \pmb{$\pm$ 0.149}&\small \pmb{$\pm$ 0.227}&\small \pmb{$\pm$ 0.175}&\small \pmb{$\pm$ 0.112}&\small \pmb{$\pm$ 0.176}&\small \pmb{$\pm$ 0.067}  \\
   	\hline
    \multirow{2}{*}{NELL995} & 55.1&17.3&14.2&41.9&51.1&26.9&18.3&16.5&12.5&28.2 \\
    &\small \pmb{$\pm$ 0.121}&\small \pmb{$\pm$ 0.185}&\small \pmb{$\pm$ 0.204}&\small \pmb{$\pm$ 0.134}&\small \pmb{$\pm$ 0.107}&\small \pmb{$\pm$ 0.125}&\small \pmb{$\pm$ 0.117}&\small \pmb{$\pm$ 0.114}&\small \pmb{$\pm$ 0.185}&\small \pmb{$\pm$ 0.062}  \\	 
    \bottomrule
  \end{tabular}
  \end{adjustbox}
  \label{tableerro1}
\end{table*}
We have run GammaE 20 times with different random seeds, and obtain mean values and standard deviations of GammaE's MRR results on EPFO and negation queries. Table~\ref{tableerro1} shows that mean values and standard deviations of GammaE's MRR results on EPFO queries. Table~\ref{tableerro2} shows mean values and standard deviations of GammaE's MRR results on negation queries. Since standard deviations of GammaE range from $\pm 0.049\%$ to $\pm 0.227\%$, the results indicate that GammaE has good robustness.

\begin{table*}[ht]
  \caption{GammaE: MRR mean values and standard deviations (\%) on answering graph queries with negation.}
  \centering
  \begin{tabular}{lcccccc}
    \toprule
    \textbf{Dataset} & \textbf{2in} & \textbf{3in} & \textbf{inp} & \textbf{pin} & \textbf{pni} & \textbf{avg} \\
    \hline
    \multirow{2}{*}{FB15k} & 20.1& 20.5& 13.5& 11.8& 17.1& 16.6\\
    &\small \pmb{$\pm$ 0.145}&\small \pmb{$\pm$ 0.175}&\small \pmb{$\pm$ 0.049}&\small \pmb{$\pm$ 0.201}&\small \pmb{$\pm$ 0.153}&\small \pmb{$\pm$0.074}\\
   	\hline
   	\multirow{2}{*}{FB15K-237} & 6.7& 9.4 & 8.6&4.8&4.4&6.78\\
   	&\small \pmb{$\pm$ 0.084}&\small \pmb{$\pm$ 0.069}&\small \pmb{$\pm$0.154}&\small \pmb{$\pm$ 0.094}&\small \pmb{$\pm$0.107}&\small \pmb{$\pm$ 0.056}\\
   	\hline
    \multirow{2}{*}{NELL995} & 6.3&8.7&11.4&4.0&4.5&6.98\\	
    &\small \pmb{$\pm$ 0.074}&\small \pmb{$\pm$ 0.142}&\small \pmb{$\pm$ 0.182}&\small \pmb{$\pm$ 0.063}&\small \pmb{$\pm$ 0.074}&\small \pmb{$\pm$ 0.059}\\ 
    \bottomrule
  \end{tabular}
  \label{tableerro2}
\end{table*}

\subsection{Computational Cost}
\label{sec:appendixcct}
To evaluate the training speed, we calculated the average time per 100 training steps. We ran all models with the same number of embedding parameters on a Tesla V100. The results are shown Table~\ref{tablecct1}. The computational cost of our GammaE is close to that of GQE, and less than those of others.

\begin{table}[ht]
  \caption{Computational costs of GammaE and the baselines. }
  \centering
  \begin{tabular}{lc}
    \toprule
    \textbf{Models} &\textbf{Running Time per 100 steps} \\
    \hline
    GQE & 12s \\
    Q2B & 15s \\
    BETAE & 24s \\  
    ConE & 18s \\
    GammaE & 15s \\     
    \bottomrule
  \end{tabular}
  \label{tablecct1}
\end{table}

The time cost of GammE is close to that of GQE, and less than those of others.

\begin{table*}[!t]
	
  \caption{Pearson correlation between learned embeddings and the number of answers on FB15K, FB15K-237, and NELL995.}
  \centering
  \begin{adjustbox}{width=\textwidth}
  \begin{tabular}{lllllllllllllll}
    \toprule
    \textbf{Dataset} &\textbf{Model} & \textbf{1p} & \textbf{2p} & \textbf{3p} & \textbf{2i} & \textbf{3i} & \textbf{pi} & \textbf{ip} & \textbf{2in} & \textbf{3in} & \textbf{inp} & \textbf{pin}& \textbf{pni} & \textbf{avg} \\
    \hline
    \multirow{3}{*}{FB15k} & BETAE & 0.22 & 0.36 & 0.38 & 0.39 & 0.30 & 0.31 & 0.31& 0.44& 0.41 & 0.34& 0.36& 0.44 & 0.36\\
    		 & ConE & 0.33 & 0.53 & 0.59 & 0.5 &0.45& 0.37 & 0.42& 0.65& 0.55& 0.50&0.52&0.64 & 0.50\\ 
    		 & GammaE& \pmb{0.39}& \pmb{0.63}& \pmb{0.65}& \pmb{0.55}& \pmb{0.53}& \pmb{0.43}& \pmb{0.53}& \pmb{0.65}& \pmb{0.61} &\pmb{0.55} & \pmb{0.59} & \pmb{0.65} & \pmb{0.56}\\
   	\hline
   	\multirow{3}{*}{FB15K-237} & BETAE & 0.23 & 0.37& 0.45& 0.36& 0.31& 0.32& 0.33& 0.46& 0.41 & 0.39& 0.36 & 0.48 &0.37 \\
    		 & ConE & 0.40& 0.52& 0.61& 0.67&0.69& 0.47& 0.49& 0.71& 0.66& 0.53&0.47&\pmb{0.72} &0.58  \\ 
    		 & GammaE& \pmb{0.49}& \pmb{0.60}& \pmb{0.63}& \pmb{0.69}& \pmb{0.69}& \pmb{0.55}& \pmb{0.54}&\pmb{0.72} &\pmb{0.70}& \pmb{0.59} & \pmb{0.52} & 0.69& \pmb{0.62} \\
   	\hline
    \multirow{3}{*}{NELL995} & BETAE & 0.24 & 0.40& 0.43& 0.40& 0.39& 0.40& 0.40& 0.52& 0.51& 0.26& 0.35& 0.46 & 0.40 \\
    		 & ConE & 0.48 & 0.45& 0.49 & \pmb{0.72} & 0.68 & 0.52 & 0.39 & \pmb{0.74}& 0.66& 0.38&0.34&\pmb{0.69} & 0.55  \\ 
    		 & GammaE& \pmb{0.53}& \pmb{0.51}&\pmb{0.51} & 0.70& \pmb{0.69}&\pmb{0.59} & \pmb{0.47}& 0.71 & \pmb{0.69}& \pmb{0.43} & \pmb{0.42} & \pmb{0.69} &\pmb{0.58}\\	 
    \bottomrule
  \end{tabular}
  \end{adjustbox}
  \label{tablec3111}
\end{table*} 

\subsection{Pearson Correlation Coefficient}
\label{sec:appendixpcc}

Pearson Correlation Coefficient measures the linear correlation of the two variables. Table~\ref{tablec3111} shows GammaE outperforms current models, and achieves up to 12.0\%, 6.89\%, 5.45\% than ConE on FB15k, FB15k-237, and NELL995.

\end{document}